\documentclass[final]{cvpr}

\usepackage{times}
\usepackage{epsfig}
\usepackage{graphicx}
\usepackage{amsmath}
\usepackage{amssymb}
\usepackage{multirow}
\usepackage{algorithm, algorithmic}
\usepackage{url}

\usepackage[pagebackref=true,breaklinks=true,colorlinks,bookmarks=false]{hyperref}

\def\cvprPaperID{4042} 
\def\confYear{CVPR 2021}

\begin{document}

\title{Rethinking Graph Neural Architecture Search from Message-passing}
\author{
Shaofei Cai$^{1,2}$, Liang Li$^{1}$\thanks{Corresponding author.}, Jincan Deng$^{1,2}$, Beichen Zhang$^{1,2}$, Zheng-Jun Zha$^{3}$, Li Su$^{2}$, Qingming Huang$^{1,2,4}$ \\
$^{1}$Key Lab of Intell. Info. Process., Inst. of Comput. Tech., CAS, Beijing, China \\
$^{2}$University of Chinese Academy of Sciences, Beijing, China \\
$^{3}$University of Science and Technology of China, China, $^{4}$Peng Cheng Laboratory, Shenzhen, China \\
{\tt \small \{shaofei.cai,jincan.deng,beichen.zhang\}@vipl.ict.ac.cn,liang.li@ict.ac.cn,zhazj@ustc.edu.cn,} \\
{\tt \small \{suli,qmhuang\}@ucas.ac.cn}
}


\maketitle

\begin{abstract}
Graph neural networks (GNNs) emerged recently as a standard toolkit for learning from data on graphs. 
Current GNN designing works depend on immense human expertise to explore different message-passing mechanisms, and require manual enumeration to determine the proper message-passing depth. 
Inspired by the strong searching capability of neural architecture search (NAS) in CNN, this paper proposes Graph Neural Architecture Search (GNAS) with novel-designed search space. The GNAS can automatically learn better architecture with the optimal depth of message passing on the graph. Specifically, we design Graph Neural Architecture Paradigm (GAP) with tree-topology computation procedure and two types of fine-grained atomic operations (feature filtering \& neighbor aggregation) from message-passing mechanism to construct powerful graph network search space. Feature filtering performs adaptive feature selection, and neighbor aggregation captures structural information and calculates neighbors’ statistics. Experiments show that our GNAS can search for better GNNs with multiple message-passing mechanisms and optimal message-passing depth. The searched network achieves remarkable improvement over state-of-the-art manual designed and search-based GNNs on five large-scale datasets at three classical graph tasks. Codes can be found at \url{https://github.com/phython96/GNAS-MP}.


\end{abstract}

\vspace{0.2cm}
\section{Introduction}
Neural architecture search automatically designs effective neural networks and has achieved remarkable performance beyond manually designed networks. Most works focus on searching CNN and RNN networks for vision and language tasks~\cite{6185680, Liu_2019_ICCV, 8695120, zha2020adversarial}, including multi-label object recognition~\cite{li2018attentive, wang2020pv}, detection~\cite{ghiasi2019fpn}, and sequence prediction~\cite{moser2020dartsrenet}. 
Recently, benefiting from the powerful reasoning capability, GNN has attracted much attention from researchers. It has become the standard toolkit for analyzing complex graph-structure data. In this paper, we introduce graph neural architecture search for improving GNNs' reasoning capability. 

The core of GNN is the message-passing mechanism on the graph, which aggregates neighbors' information and updates center node representations. 
The common message-passing mechanisms can be divided into two classes: (1) \emph{isotropic mechanism} (e.g. GCN~\cite{kipf2016semi}, GraphSage~\cite{hamilton2017inductive}) treats every ``edge direction" equally in node update equation. 
(2) \emph{anisotropic mechanism} (e.g. GAT~\cite{velickovic2018graph}, GatedGCN~\cite{bresson2017residual}) assigns weight for every edge according to joint representations of adjacent nodes. 
For example, GAT and GatedGCN compute edge weights based on sparse attention and dense attention mechanisms, respectively~\cite{dwivedi2020benchmarking}. 
Each mechanism has its characteristics of information transmission. Current GNNs are usually stacked to multiple layers with the same message-passing mechanism to capture long-range node dependencies. 
An onefold message-passing mechanism limits the reasoning power of graph networks. 
However, manually designing GNNs with multiple message-passing mechanisms requires immense human expertise. 

\begin{figure}
    \centering
    \includegraphics[scale=0.45, trim = 10 0 0 0,clip]{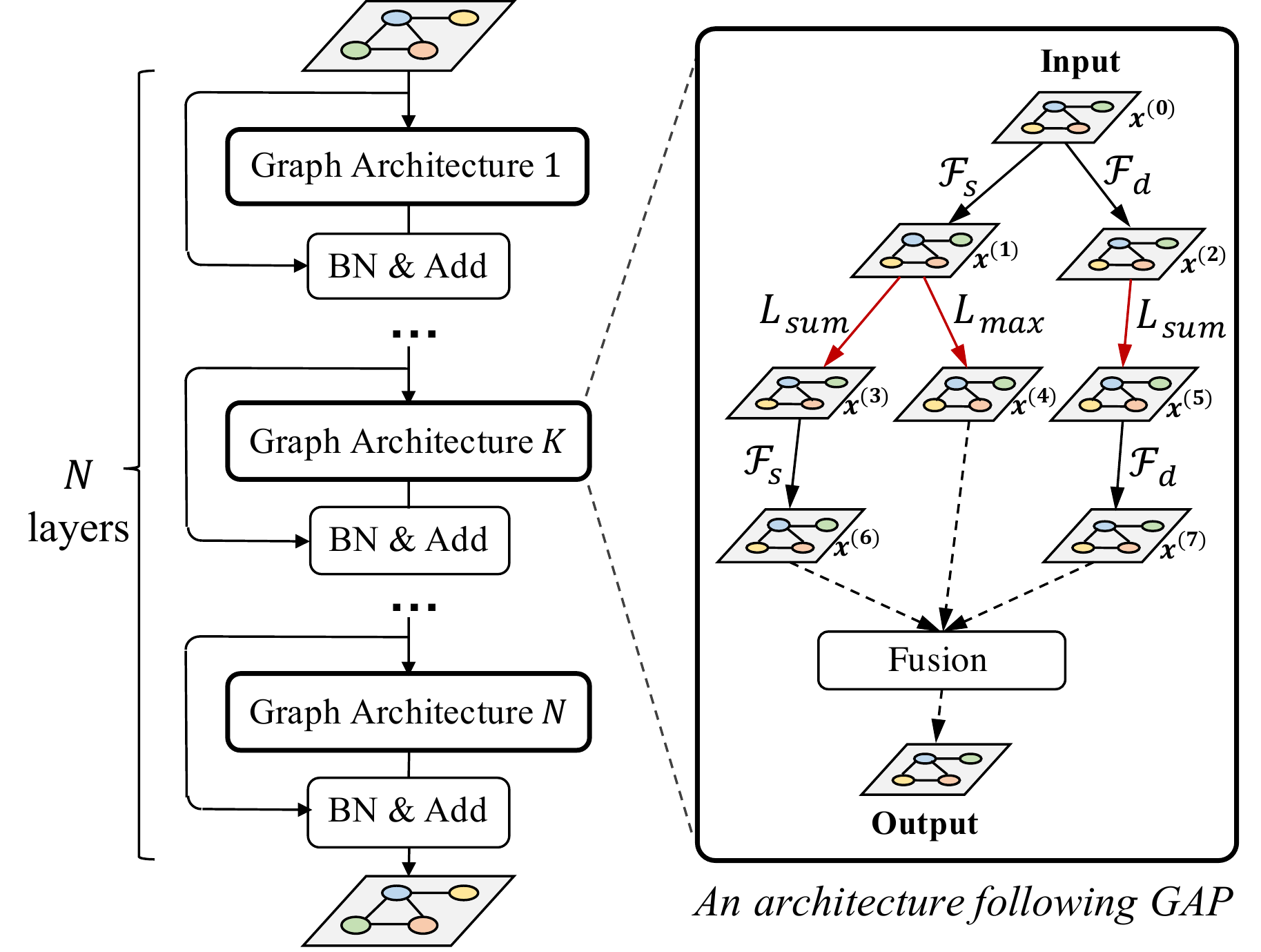}
    \vspace{0.5cm}
    \caption{An overview of graph neural network with $N$ layers searched by GNAS where each graph architecture layer follows GAP. ``BN \& Add'' module first applies batch normalization to the output of the last graph architecture layer and then adds the input of that. ``Fusion'' module (such as sum pooling) fuses the computation tree branches to calculate the final output of each graph architecture layer. $\mathcal{F}_s,\mathcal{F}_d$ are feature filtering operations and $L_{sum},L_{max},L_{mean}$ are neighbor aggregation operations. }
    \label{fig_pipeline}
\end{figure}

Another critical problem for GNN is determining the number of graph convolution layers, that is, the depth of message-passing. 
Different from CNN, recent works~\cite{li2018deeper,rong2019dropedge,Wu_2020} show that GNN's reasoning capability degrades as the network goes too deep. 
This results from that the representations of adjacent nodes become closer to each other after each graph convolution. 
In theory, with an extreme depth, all nodes' representations will converge to a stationary point. 
Further, the network depth is dataset-relevant. Specifically, it depends on the diameter of the graph in the specific dataset. 
In order to find the optimal network depth, current works usually use enumeration with the high computational cost. 
NAS has achieved great success by searching for efficient operations in vast search space and discovering excellent representation network. Motivated by this, we explore NAS for GNN to solve the above problems. 
Search space and search strategy are the most essential components in NAS. 
The search space defines which architectures can be represented in principle. The search strategy details how to explore the search space, which is mainly classified into reinforcement learning (RL)~\cite{baker2016designing, zoph2016neural, zoph2018learning}, evolutionary algorithms (EA)~\cite{liu2017hierarchical, real2019regularized, real2017large} and gradient-based (GB)~\cite{li2020sgas, liu2018darts, xu2019pc, zela2019understanding} methods. 
However, traditional NAS~\cite{zhou2020posterior} methods cannot directly process graph-structure data because atomic operations (such as convolutions, pooling) in search space come from the CNN and RNN domains. 
Recently, researchers~\cite{nunes2020neural, zhou2019auto} use existing GNNs (GCN~\cite{kipf2016semi}, GAT~\cite{velickovic2018graph}, etc.) and hyper parameters (the head number of GAT, etc.) as atomic operations for searching. 
It's essentially a kind ensemble and fine-tuning of the existing GNNs, instead of deriving a new GNN from the message-passing mechanism. The coarse-grained operations~(existing GNNs) cause redundant computation and limit the searching upper bound for network reasoning capability. 
In this paper, we propose Graph Neural Architecture Search (GNAS) with novel-designed search space and gradient-based search strategy to automatically learn better architecture with an optimal depth of message-passing on the graph. 
To raise the searching upper bound for higher performance, we deconstruct GNN from the message-passing mechanism and design Graph Neural Architecture Paradigm (GAP). GAP introduces a tree-topology computation procedure with two types of fine-grained atomic operations to construct graph neural networks: (1) \emph{feature filtering} plays a role in adaptive feature selection using gating mechanism, (2) \emph{neighbor aggregation} captures structural information via sum operation and calculates neighborhood statistics with max and mean operations. 
Theoretically, recent popular GNNs (GCN~\cite{kipf2016semi}, GIN~\cite{xu2018how}, GraphSage~\cite{hamilton2017inductive}, GAT~\cite{velickovic2018graph}, GatedGCN~\cite{bresson2017residual}, etc.) can be approximated as the special case of GAP. 
Following the paradigm, we design a three-level search space and adopt a gradient-based search strategy for architecture optimization. 
Figure~\ref{fig_pipeline} shows an example of a graph neural network searched by GNAS. 
Experiment results show that our GNAS can search better graph networks than state-of-the-art manually designed and search-based GNNs on five large-scale datasets at three classical graph tasks. 
Moreover, as a significant finding, experiments demonstrate that GNAS can search the optimal depth rather than predefined depth of message-passing. 

Our contributions can be summarized as follows:
\begin{itemize}
    \item We propose a novel Graph Neural Architecture Paradigm (GAP) with a tree-topology computation procedure and two types of fine-grained atomic operations to construct powerful graph neural networks. 
    \item Following the GAP and gradient-based search strategy, we propose Graph Neural Architecture Search to automatically learn better GNN architecture with an optimal depth of message-passing on the graph. 
    \item We conduct extensive experiments on five datasets at three classical tasks, and the results show the superiority of our GNAS over SOTA manually designed and search-based GNNs. 
\end{itemize}

\section {Graph Neural Architecture Paradigm}
In this section, we first detail the topology of computation procedure for graph architecture. 
Second, we introduce two kinds of operations to construct powerful graph architecture space: feature filtering and neighbor aggregation. We then describe how to calculate the final output of architecture. Finally, based on this paradigm, we formulize the approximation of the existing GNNs including GCN, GIN, GraphSAGE, GAT and GatedGCN from GAP view. 

\subsection{Architecture} \label{sec_GNA}

GAP defines the topology of graph neural architecture as a directed tree. Each node $x^{(i)}$ is a latent representation ($x^{(0)}$ denotes node embeddings of input graph) and each directed edge $(i,j)$ is associated with one operation that transforms $x^{(i)}$ to $x^{(j)}$. 
From the message-passing mechanism perspective, feature filtering is responsible for re-scaling message, and neighbor aggregation is in charge of passing the message on the graph. 

\noindent
\textbf{Feature filtering.} This kind of operation plays a role in adaptive feature selection for each node by using a gating mechanism to control the information flow. We design the sparse filter $\mathcal{F}_s(\cdot)$ and dense filter $\mathcal{F}_d(\cdot)$ to perform coarse-grained and fine-grained re-scaling, respectively. This computation procedure can be formulated as
\begin{equation}
    \mathcal{F}_{s}(\textbf{H}) = \textbf{QH},
\end{equation}
\begin{equation}
   \mathcal{F}_{d}(\textbf{H}) = \textbf{Z}\odot \textbf{H},
\end{equation}
where $\odot$ denotes hardmard product, $\textbf{Q} \in \mathbb{R}^{n\times n}$ and $\textbf{Z} \in \mathbb{R}^{n\times d}$ denote the re-scaling matrix to reweight node embeddings $\textbf{H}\in \mathbb{R}^{n \times d}$. 
Here, we introduce $\textbf{H}_{in}$ for jointly computing re-scaling factors, described as
\begin{equation}
    \textbf{Q} = diag(\mathcal{M}_{Q}([\textbf{H}, \textbf{H}_{in}])),
\end{equation}
\begin{equation}
    \textbf{Z} = \mathcal{M}_{Z}([\textbf{H}, \textbf{H}_{in}]),
\end{equation}
where $\mathcal{M}_Q(\cdot), \mathcal{M}_Z(\cdot)$ denote $\mathbb{R}^{2\times d}$-to-$\mathbb{R}$ and $\mathbb{R}^{2\times d}$-to-$\mathbb{R}^{d}$ multilayer perceptron, respectively, $diag(\cdot)$ converts the vector into diagonal matrix. Inspired by the gating mechanism, we used $\sigma(fc(\cdot))$ to simplify $\mathcal{M}_{Q}(\cdot)$ and $\mathcal{M}_{Z}(\cdot)$, where $fc(\cdot)$ denotes fully-connected layer, $\sigma(\cdot)$ denotes the sigmoid function. Besides, we design an identity filter to help incorporate each node’s own features, which is similar to a residual connection and is described as
\begin{equation}
   \mathcal{I}(\textbf{H}) = \textbf{H}.
\end{equation}

\begin{figure}[tbp]   
    \centering
    \includegraphics[scale=0.60, trim = 0 0 8 0,clip]{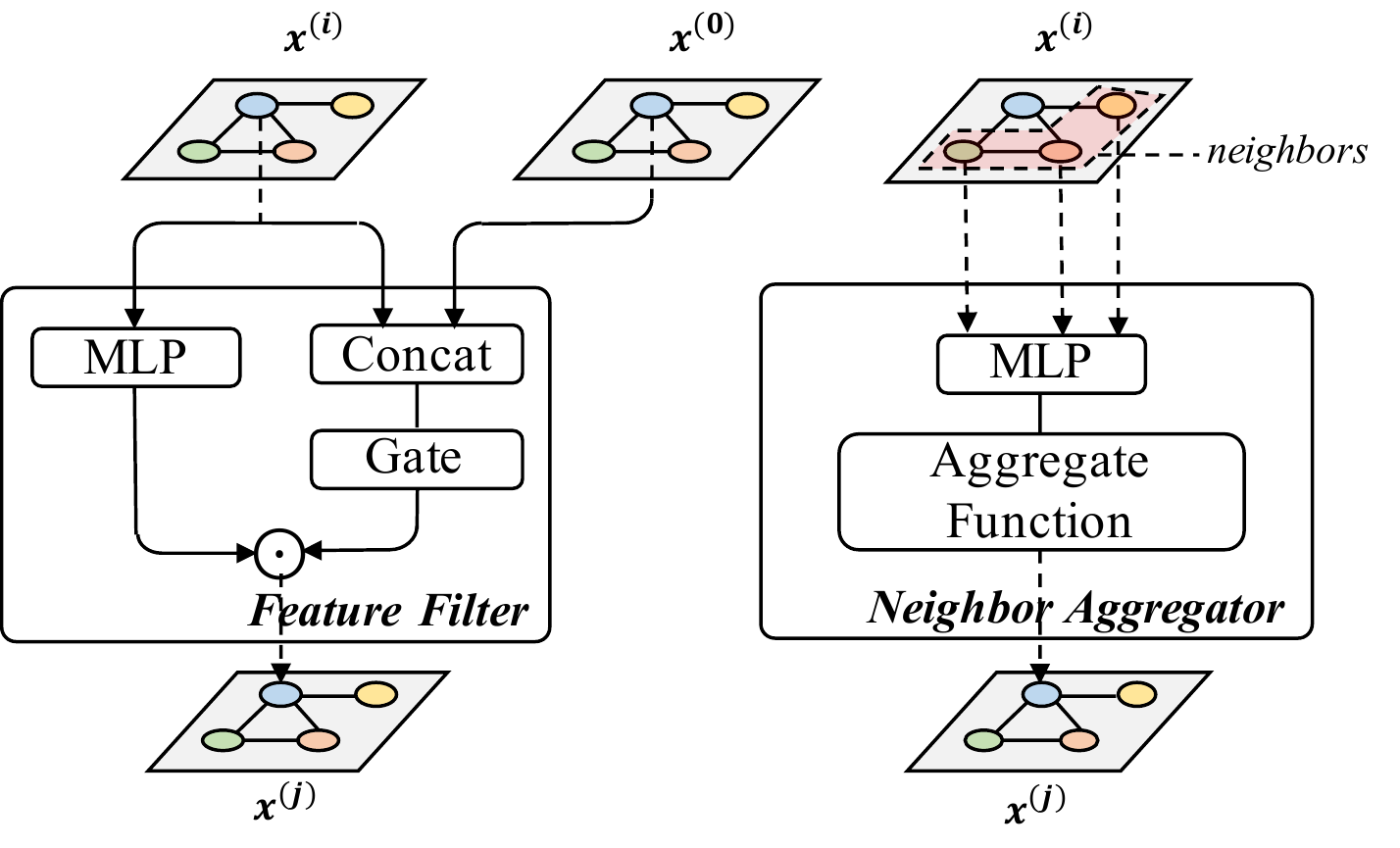}
    \caption{
    The computation procedure of fine-grained atomic operations: feature filtering and neighbor aggregation operations. $x^{(0)}$ is the latent input representation of graph architecture. 
    ``Gate'' is the module to compute the adaptive scaling factor. 
    ``Concat'' denotes the concatenation operation. ``Aggregate function'' denotes a continuous function of multisets (e.g. sum, mean, max) .}

    \label{fig_operations}
 \end{figure}

\noindent 
\textbf{Neighbor aggregation.} Neighbor aggregation captures structural information and calculates neighborhood statistics. We define the aggregators as continuous functions of multisets which aggregate information on neighbor nodes, such as \emph{max}, \emph{mean} and \emph{sum}. 
Different aggregators capture different types of information. Work~\cite{xu2018how} demonstrates that \emph{sum} aggregator does well in capturing structural information while \emph{max} aggregator identifies representative elements or the ``skeleton'' and is robust to noise and outliers. Additionally, \emph{mean} aggregator extracts statistics from the input message, and allows the centre node to understand the distribution of messages it receives. 
Considering that the aggregators are complementary, GAP jointly uses multiple aggregators to enhance the expressive power of GNN. 

\begin{figure*}[htbp] 
    \centering
    \includegraphics[scale=0.46]{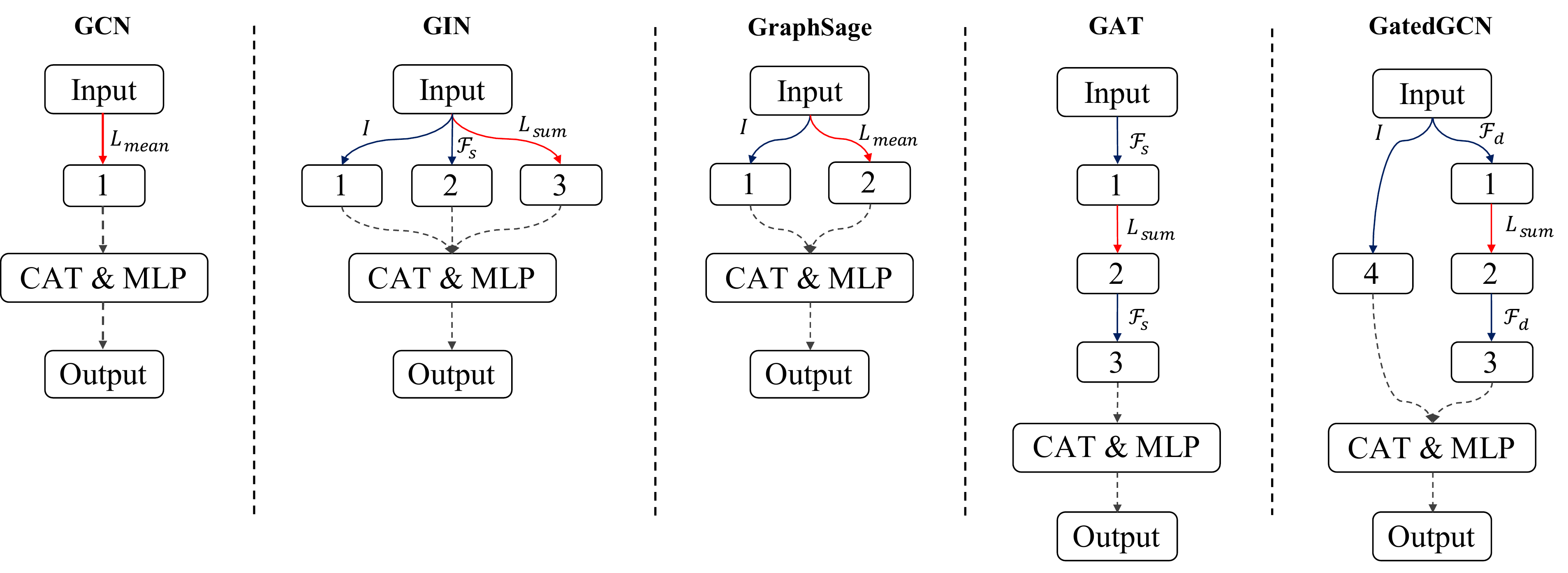}
    \caption{Illustrations of approximating manual designed GNNs. 
    $\mathcal{F}_s,\mathcal{F}_d$ are feature filtering operations and $L_{sum},L_{max},L_{mean}$ are neighbor aggregation operations. $I$ denotes identity operation. 
    ``CAT \& MLP'' module first concatenates the branches of the computation tree and then uses a multilayer perceptron to calculate the output. }
    \label{fig_approx}
 \end{figure*}

\begin{table}[] 
    \small
    \setlength{\tabcolsep}{2.0 mm}
    \renewcommand\arraystretch{1.5}
    \begin{tabular}{c|c}
    \hline
    GNNs      & Approximation Formula\\ \hline \hline
    GCN       &$\textbf{H}_{out} \approx \mathcal{M}(L_{mean}(\textbf{H}_{in}))$\\ \hline
    GIN       &$\textbf{H}_{out} \approx \mathcal{M}([\mathcal{I}(\textbf{H}_{in}) || \mathcal{F}_s(\textbf{H}_{in})||L_{sum}(\textbf{H}_{in})])$\\ \hline
    GraphSage &$\textbf{H}_{out} \approx \mathcal{M}([\mathcal{I}(\textbf{H}_{in}) \parallel L_{mean}(\textbf{H}_{in})])$\\ \hline
    GAT       &$\textbf{H}_{out} \approx \mathcal{M}(\mathcal{F}_s(L_{sum}(\mathcal{F}_s(\textbf{H}_{in}))))$\\ \hline
    GatedGCN  &$\textbf{H}_{out} \approx \mathcal{M}([\mathcal{I}(\textbf{H}_{in})\parallel \mathcal{F}_d(L_{sum}(\mathcal{F}_d(\textbf{H}_{in})))])$\\ \hline
    \end{tabular}
    \vspace{0.5em}
    \caption{Approximation formula for manually designed GNN networks (e.g. GCN, GIN, GraphSage, GAT and GatedGCN) from GAP view. }
    \label{tab_approx}
 \end{table}

\noindent
\textbf{Output of architecture.} All leaf nodes in the computation tree are taken into account when calculating the output. Any continuous function of multisets can be used to fuse the hidden embeddings. Specifically, we concatenate the hidden embeddings and feed it into the multilayer perceptron to calculate the final output, described as
\begin{equation}
    \textbf{H}_{out} = \mathcal{M}(Concat(\{\textbf{H}|\textbf{H} \in \mathcal{A}\})),
\end{equation} 
where $\mathcal{A}$ denotes the set of leaf nodes in the computation tree, $\mathcal{M}(\cdot)$ is multilayer perceptron.

Notably, in GAP, each root-to-leaf path contains at most one neighbor aggregation operation, which means each node can only access its first-order neighbor information. This allows the architecture to be compared to other GNN. More importantly, we can control the size of the neighborhood receptive field by simply changing network depth.

Here, we discuss the role of jointly using feature filtering operations and neighbor aggregation operations from the message-passing mechanism perspective. 
In message-passing, the source node sends the message, and the destination node receives the message. 
The feature filtering before the neighbor aggregation adaptively re-scales the message to send to the neighbors. Similarly, the feature filtering after the neighbor aggregation adaptively retains critical messages received from the neighbors. The flexible combination of the two kind of operations helps explore rich message-passing model. 


\subsection{GAP View for Traditional GNNs}
GAP defines a kind of GNN designing paradigm, by which most traditional GNNs~(e.g., GCN, GIN, GraphSage, GAT, GatedGCN) can be represented. 
These GNNs perform nested operations in Section \ref{sec_GNA} to compute latent representations and concatenate them into a multilayer perceptron for output. 
All formulation approximation results are shown in Table \ref{tab_approx} and illustrated in Figure \ref{fig_approx}. The detailed derivation can be found in the appendix. 

\begin{figure}[tbp]
    \centering
    \includegraphics[scale=0.30]{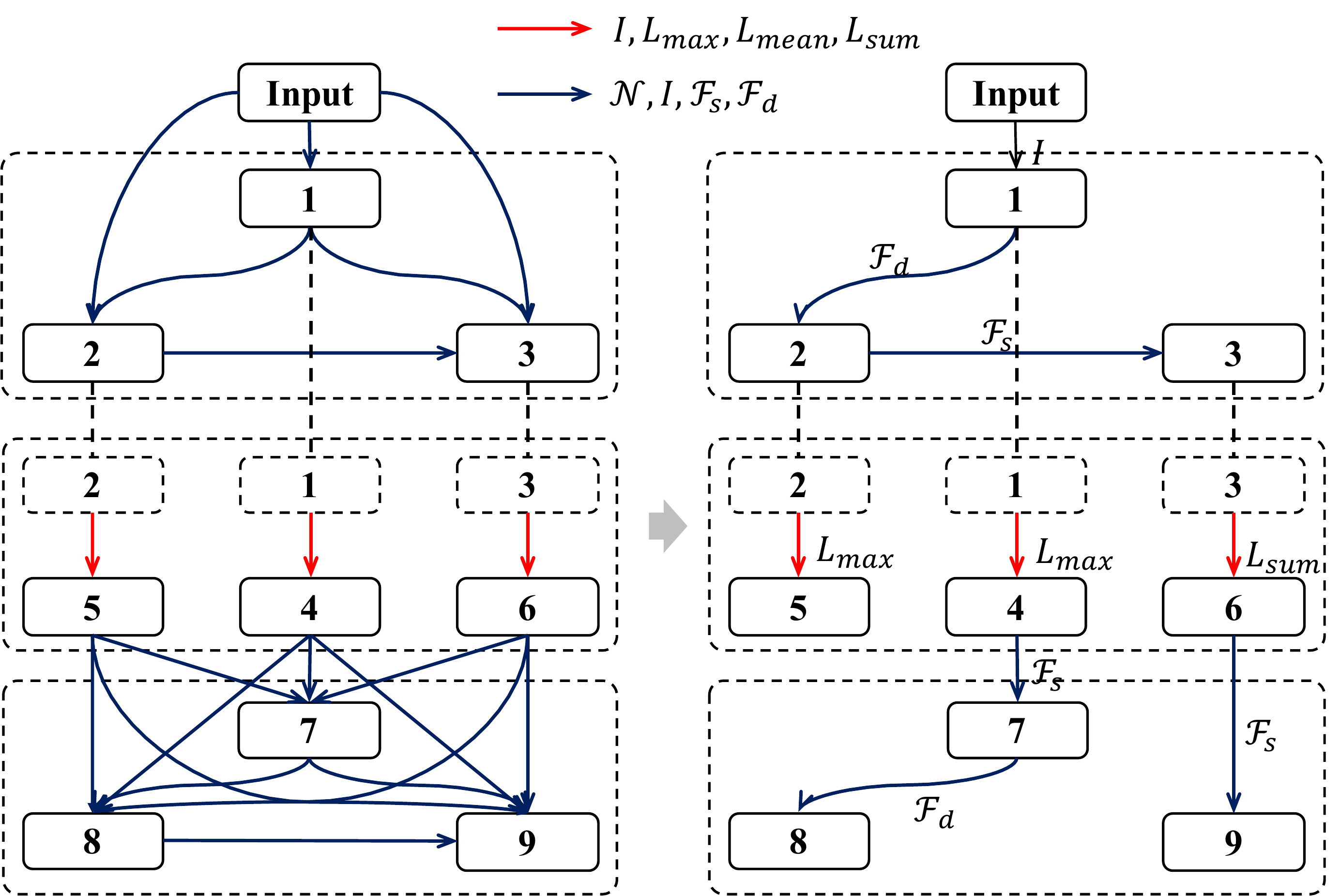}
    \caption{The illustration of deriving a discrete graph neural architecture from the search space (left figure) by cutting out edges and selecting important operations. 
    Blue and red edges represent the feature filtering and neighbor aggregation operation, respectively. }
    \label{fig_searchspace}
 \end{figure}

\section{Graph Neural Architecture Search}
Following graph neural architecture paradigm (GAP), we design a three-level search space. We then introduce DARTS~\cite{liu2018darts} algorithm to perform continuous relaxation for search space and joint optimize the architecture and its weights. After optimization, we show how to derive a discrete sub-architecture from super-architecture. Finally, we detail how GNAS searches the optimal depth of message-passing for each specific dataset. 

\subsection{Search Space}

We search for computation cells as the building blocks and stack them for the final model. 
Considering that each root-to-leaf path contains at most one neighbor aggregation operation, we propose a three-level search space~(illustrated in Figure \ref{fig_searchspace}), where only the second level can use neighbor aggregation operations. 

The first level is a directed acyclic graph consisting of an ordered sequence of $N$ nodes. Each node $x^{(i)}$ is a hidden embedding and each directed edge $(i, j)$ is associated with a feature filtering operation $o_{\mathcal{F}}^{(i,j)}$ that transforms $x^{(i)}$. We also use a special $zero$ operation to indicate a lack of connection between two nodes, which is denoted as $\mathcal{N}$. Each intermediate node is computed based on all its predecessors: 
\begin{equation}
   x^{(j)} = \sum_{0\le i<j}o_{\mathcal{F}}^{(i,j)}(x^{(i)}).
\end{equation}
where $1 \le j \le N$, $0 \le i < j$, $x^{(0)}$ denotes input (root) embedding. Let $\mathcal{O}_{\mathcal{F}} = \{\mathcal{N},\mathcal{I}, \mathcal{F}_{s}, \mathcal{F}_{d}\}$ be a set of candidate feature filtering operations, where $o_{\mathcal{F}}^{(i,j)} \in \mathcal{O}_{\mathcal{F}}$. 

The second level consists of an ordered sequence of $N$ nodes and exactly $N$ edges. The nodes are numbered from $N+1$ to $2N$. The $i$-th edge $(i, N+i)$ connects the node $x^{(i)}$ in first level and node $x^{(N+i)}$ in second level. It is associated with a neighbor aggregation operation $o_{L}^{(i,N+i)}$ that transforms $x^{(i)}$. Specifically, we add identity operation $\mathcal{I}$ to accommodate situations that do not require neighborhood information. Each intermediate node in the second level is computed based on its predecessor in the first level:
\begin{equation}
   x^{(N+i)} = o_{L}^{(i,N+i)}(x^{(i)})
\end{equation}
where $1 \le i \le N$. Let $\mathcal{O}_{L} = \{\mathcal{I}, L_{sum}, L_{mean}, L_{max}\}$ be a set of candidate neighbor aggregation operations, where $o_{L}^{(i,N+i)} \in \mathcal{O}_{L}$. Note that, there is no connection between any paired nodes in second level. 

The third level is also a directed acyclic graph consisting of an ordered sequence of $M$ nodes. Unlike the first level with only one input node, the third level takes $N$ nodes of the second level as input. The edge is associated with feature filtering operation $o_{\mathcal{F}}^{(i,2N+j)}$ same as the first level. Each intermediate node is computed based on all its predecessors:
\begin{equation}
   x^{(2N+j)} = \sum_{N+1 \le i < 2N+j}o_{\mathcal{F}}^{(i,2N+j)}(x^{(i)}),
\end{equation}
where $1 \le j \le M$. 

\begin{figure}[tbp]
    \centering
    \includegraphics[scale=0.50]{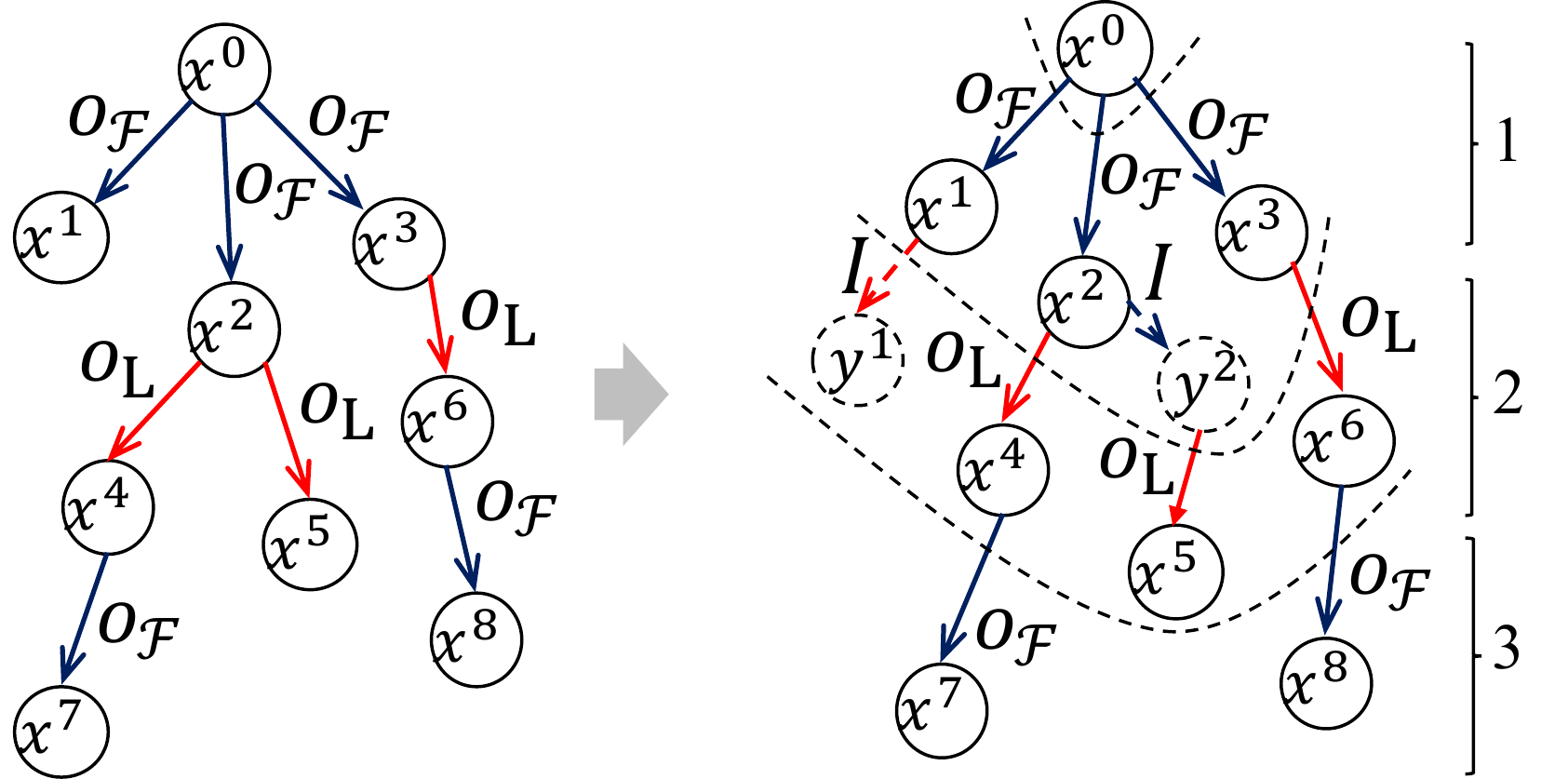}
    \caption{For an arbitrary graph neural architecture that follows GAP (left figure), we can derive an equivalent variant from the three-level search space (right figure).}
    \label{fig_treetothreelevel}
 \end{figure}

It is easy to prove that \emph{any graph neural architecture that follows GAP can be represented into three-level search space by adding identity operation $\mathcal{I}$ and new nodes}. 
For the particular situation~(Figure~\ref{fig_treetothreelevel}) that a single node emits multiple edges with neighbor aggregation operation, we propose the following solution. 
For each node $x^{(i)}$ that has $e$ ($e > 1$) outgoing edges associated with neighbor aggregation operation, we add $(e-1)$ nodes connected to $x^{(i)}$ using identity operation. 
We then pick the $(e-1)$ successor nodes of $x^{(i)}$ and connect them to the above new $(e-1)$ nodes one by one. 
For each root-to-leaf path without neighbor operation, we add a new node to the end of the path with identity connection method. 
The above rules guarantee that all root-to-leaf paths contain exactly one neighbor aggregation operation. 
Finally, the ancestor nodes of all edges associated with neighbor aggregation operation are divided into the first level. The outgoing nodes of these edges are divided into the second level. All other nodes are divided into the third level. 
An example is illustrated in Figure \ref{fig_treetothreelevel}. 
This neat proof shows that our three-level search space can construct a vast space of GNN.

\subsection{Continuous Relaxation and Optimization}

We use the same method as DARTS to make the search space continuous. We relax the categorical choice of a particular operation to a softmax over all possible operations:
\begin{equation}
   \bar{o}^{(i,j)}(x) = \sum_{o \in \mathcal{O}}softmax(\alpha_{o}^{(i,j)})o(x)
\end{equation}
where the operation mixing weights for a pair of nodes $(i, j)$ are parameterized by a vector $\alpha^{(i,j)}$ of dimension $|\mathcal{O}|$. $\mathcal{O}$ can be either $\mathcal{O}_{\mathcal{F}}$ or $\mathcal{O}_{L}$. The task of architecture search then reduces to learning a set of continuous variables $\alpha = \{\alpha^{(i,j)}\}$. 

After relaxation, following DARTS~\cite{liu2018darts}, we jointly learn architecture $\alpha$ and the weights $w$ within all the mixed operations. $\alpha$ and $w$ can be efficiently optimized using differentiable methods. 
At the end of search, a discrete architecture can be obtained by replacing each mixed operation $\bar{o}^{(i,j)}$ with the most likely operation, i.e. 
\begin{equation}
   o^{(i,j)} = argmax_{o \in \mathcal{O}} (\alpha_{o}^{(i,j)}). 
\end{equation}
For each edge, we only retain the strongest non-zero operation. For each node, we only retain the strongest edge from all of its incoming edges, where the strength of an edge is the strength of its strongest operation. 

\begin{algorithm}[t]
	\renewcommand{\algorithmicrequire}{\textbf{Input:}}
	\renewcommand{\algorithmicensure}{\textbf{Output:}}
	\caption{Search Efficient GNN with Optimal message-passing Depth}
	\label{alg:algorithm1}
	\begin{algorithmic}[1]
		\REQUIRE dataset $\mathcal{S}$
        \ENSURE graph neural network $\mathcal{N}$
        \STATE Initialize $\mathcal{D}_{o}$ as half of average graph diameter of $\mathcal{S}$
        \REPEAT 
        \STATE Initialize $\mathcal{N}_{s}$ as a search network with $\mathcal{D}_{o}$-layer graph neural architecture
        \STATE Optimize the architectures of $\mathcal{N}_{s}$ with GNAS on $\mathcal{S}$
        \STATE Derive a discrete sub-network of $\mathcal{N}_{d}$ from $\mathcal{N}_{s}$  
        \STATE $\mathcal{D}_{i}$ = $\mathcal{D}_{o}$ 
        \STATE Update $\mathcal{D}_{o}$ as the number of cells with at least one neighbor aggregation in $\mathcal{N}_{d}$
        \UNTIL $\mathcal{D}_{i} = \mathcal{D}_{o}$
		\STATE \textbf{return} $\mathcal{N}_{d}$
    \end{algorithmic}  
\end{algorithm}

\begin{table*}[tbp]
    \small
    \setlength{\tabcolsep}{2.6mm}
    \renewcommand\arraystretch{1.15}
    \begin{tabular}{c|c|c|c|c|c|c|c|c|c}
    \hline
    \multicolumn{2}{c|}{} & \multicolumn{8}{c }{NODE CLASSIFICATION}                                        \\ \hline
    \multicolumn{2}{c|}{} & \multicolumn{4}{c|}{PATTERN}           & \multicolumn{4}{c }{CLUSTER}           \\ \hline
    Model                                       & Depth & \#Params & Test Acc $\pm$ std & Search & Train & \#Params & Test Acc $\pm$ std & Search & Train \\ \hline \hline
    MLP                                         & 4 & 0.11M & $50.52 \pm 0.00$ & - & 0.11 hr & 0.11M & $20.94 \pm 0.00$ & - & 0.07 hr \\ 
    GCN~\cite{kipf2016semi}                     & 4 & 0.10M & $63.88 \pm 0.07$ & - & 3.51 hr & 0.10M & $53.45 \pm 2.03$ & - & 1.30 hr \\ 
    GIN~\cite{xu2018how}                        & 4 & 0.10M & $85.59 \pm 0.01$ & - & 0.40 hr & 0.10M & $58.38 \pm 0.24$ & - & 0.23 hr \\ 
    GraphSage~\cite{hamilton2017inductive}      & 4 & 0.10M & $50.52 \pm 0.00$ & - & 1.17 hr & 0.10M & $50.45 \pm 0.15$ & - & 0.97 hr \\ 
    GAT~\cite{velickovic2018graph}              & 4 & 0.11M & $75.82 \pm 1.82$ & - & 0.57 hr & 0.11M & $57.73 \pm 0.32$ & - & 0.27 hr \\ 
    GatedGCN~\cite{bresson2017residual}         & 4 & 0.10M & $84.48 \pm 0.12$ & - & 3.09 hr & 0.10M & $60.40 \pm 0.42$ & - & 2.13 hr \\ 
    MoNet~\cite{monti2017geometric}             & 4 & 0.10M & $85.48 \pm 0.04$ & - & 0.90 hr & 0.10M & $58.06 \pm 0.13$ & - & 0.52 hr \\ \hline 
    GraphNAS~\cite{gao2020graph}                & 4 & 0.48M & $85.21 \pm 0.01$ & 120 hr & 8.25 hr & 0.48M & $52.61 \pm 0.22$ & 120 hr & 9.50 hr \\ 
    \textbf{GNAS}                               & 4 & 0.35M & $\textbf{86.80} \pm \textbf{0.10}$ & 2.45 hr & 2.15 hr & 0.38M & $\textbf{62.21} \pm \textbf{0.20}$ & 2.50 hr & 1.20 hr\\ \hline
    \end{tabular}
    \vspace{0.5em}
    \caption{Results of the model searched by our GNAS in comparision with \emph{state-of-the-art} methods on node classification task, including number of parameters, accuracy, searching cost and training cost. }
    \label{tab_nc}
\end{table*}

\subsection{Optimal Depth of Message-passing}
Message-passing depth is the number of stacked graph architectures with neighbor aggregation, which determines neighborhood receptive field size. 
For traditional GNN, the message-passing depth is equal to the network depth because each layer of GNN aggregates the representations from neighbor nodes. 
Works~\cite{li2018deeper,rong2019dropedge,Wu_2020} point that when the network goes too deep, the aggregated information of the center node quickly covers the whole graph, and all features of nodes tend to converge to the same so that the nodes lose their discriminability. 
Researchers usually take the depth as a hyper parameter and determine the optimal depth by enumeration. 
This brings a huge cost of time and computation and requires a rich human experience. 
As a neural architecture search algorithm, our GNAS can not only search for efficient GNN but also learn the optimal depth of message-passing, which is detailed in Algorithm~\ref{alg:algorithm1}. 

\begin{table}[]
    \small
    \setlength{\tabcolsep}{1.0mm}
    \renewcommand\arraystretch{1.3}
    \begin{tabular}{c|c|c|c|c}
    \hline
    Task                                 & Dataset & Graphs & Nodes  & Total Nodes \\ \hline \hline
    Graph Regression                     & ZINC & 12K    & 9-37   & 277,864     \\ \hline
    \multirow{2}{*}{Node Classification} & PATTERN & 14K    & 44-188 & 1,664,491   \\ 
                                         & CLUSTER & 12K    & 41-190 & 1,406,436   \\ \hline
    \multirow{2}{*}{Graph Classification}& MNIST   & 70K    & 40-75  & 4,939,668   \\ 
                                         & CIFAR10 & 60K    & 85-150 & 7,058,005   \\ \hline
    \end{tabular}
    \vspace{0.2em}
    \caption{Statistics of datasets used to evaluate the methods.}
    \label{tab_datasets}
\end{table}

\begin{table*}[htbp]
    \small
    \setlength{\tabcolsep}{2.8 mm}
    \renewcommand\arraystretch{1.15}
    \begin{tabular}{c|c|c|c|c|c|c|c|c|c}
    \hline
    \multicolumn{2}{c|}{} & \multicolumn{8}{c}{GRAPH CLASSIFICATION}                                       \\ \hline
    \multicolumn{2}{c|}{} & \multicolumn{4}{c|}{MNIST}             & \multicolumn{4}{c}{CIFAR10}           \\ \hline
    Model       & Depth   & \#Param  & Test Acc $\pm$ std & Search & Train & \#Param  & Test Acc $\pm$ std & Search & Train   \\ \hline \hline
    MLP                                     & 4 & 0.10M & $95.34 \pm 0.14$ & - & 1.48 hr & 0.10M & $56.34 \pm 0.18$ & - & 1.53 hr \\ 
    GCN~\cite{kipf2016semi}                 & 4 & 0.10M & $90.71 \pm 0.22$ & - & 2.99 hr & 0.10M & $55.71 \pm 0.38$ & - & 4.39 hr \\ 
    GIN~\cite{xu2018how}                    & 4 & 0.10M & $96.49 \pm 0.25$ & - & 1.41 hr & 0.11M & $55.26 \pm 1.53$ & - & 2.07 hr \\ 
    GraphSage~\cite{hamilton2017inductive}  & 4 & 0.10M & $97.31 \pm 0.10$ & - & 3.13 hr & 0.10M & $65.77 \pm 0.31$ & - & 3.29 hr \\ 
    GAT~\cite{velickovic2018graph}          & 4 & 0.11M & $95.54 \pm 0.21$ & - & 1.25 hr & 0.11M & $64.22 \pm 0.46$ & - & 1.62 hr \\ 
    GatedGCN~\cite{bresson2017residual}     & 4 & 0.10M & $97.34 \pm 0.14$ & - & 3.50 hr & 0.10M & $67.31 \pm 0.31$ & - & 4.22 hr \\ 
    MoNet~\cite{monti2017geometric}         & 4 & 0.10M & $90.81 \pm 0.03$ & - & 3.82 hr & 0.10M & $54.66 \pm 0.52$ & - & 3.85 hr \\ \hline
    GraphNAS~\cite{gao2020graph}            & 4 & 0.48M & $93.80 \pm 0.10$ & 120 hr & 9.85 hr & 0.48M & $58.33 \pm 0.63$ & 120 hr & 11.2 hr \\
    \textbf{GNAS}                           & 4 & 0.39M & $\textbf{98.01} \pm \textbf{0.10}$ & 6.00 hr & 3.10 hr& 0.43M & $\textbf{70.10} \pm \textbf{0.44}$ & 7.20 hr & 3.45 hr \\ \hline
    \end{tabular}
    \vspace{0.5em}
    \caption{Results of the model searched by our GNAS in comparision with \emph{state-of-the-art} methods on graph classification task, including number of parameters, accuracy, searching cost and training cost. }
    \label{tab_gc} 
\end{table*}

\section{Experiments}

\subsection{Experimental setting}
\noindent
\textbf{Datasets}.
Recent work~\cite{dwivedi2020benchmarking} points that existing benchmarks such as Cora~\cite{mccallum2000automating}, Citeseer~\cite{getoor2005link} and TU~\cite{KKMMN2016} are too simple to distinguish the representation power of complex GNNs. Consequently, a new range of datasets across different real-world tasks is proposed in~\cite{dwivedi2020benchmarking}. To evaluate the search performance of our GNAS, we access it on these datasets of three tasks, including chemistry (ZINC~\cite{irwin2012zinc}), mathematical modeling (PATTERN~\cite{dwivedi2020benchmarking} and CLUSTER~\cite{dwivedi2020benchmarking}) and computer vision (CIFAR10~\cite{krizhevsky2009learning} and MNIST~\cite{lecun1998gradient}). ZINC~\cite{irwin2012zinc} is one popular real-world molecular dataset of 250K graphs, whose task is graph property regression, out of which we randomly select 12K for efficiency. PATTERN~\cite{dwivedi2020benchmarking} and CLUSTER~\cite{dwivedi2020benchmarking} are node classification tasks generated via Stochastic Block Models~\cite{abbe2017community}, which are used to model communities in social networks by modulating the intra- and extra-communities connections. MNIST~\cite{lecun1998gradient} and CIFAR10~\cite{krizhevsky2009learning} are classical image classification datasets converted into graphs using super-pixels~\cite{achanta2012slic} which assigns each node’s features as the super-pixel coordinates and intensity. Details about the five datasets are shown in Table \ref{tab_datasets}.

\noindent
\textbf{Searching setting.} 
In GNAS, we define the operation set $\mathcal{O}$: \emph{sum aggregator}, \emph{max aggregator}, \emph{mean aggregator}, \emph{identity}, \emph{sparse filter}, \emph{dense filter} and \emph{zero}. Every operation except zero is followed by the linear transformation function and the activation function ReLU~\cite{nair2010rectified}. In three-level search space, we set $3$ nodes at each level. For each computation cell, we concatenate all nodes in third level and pass them into FC-BN-ReLU to get the final output. Besides, in order to stabilize the gradient, additional residual connections are introduced. To carry out the architecture search, we hold out half of the training data as the validation set. A small network of 4 layers is trained using GNAS for 50 epochs, with batch size 64 (for both the training and validation set). We use momentum SGD to optimize the weights $w$, with initial learning rate $\eta_{w}=0.025$ (annealed down to zero following a cosine schedule without restart), momentum $0.9$, and weight decay $3 \times 10 ^{-4}$. We use Adam~\cite{maclaurin2015gradient} as the optimizer for $\alpha$, with initial learning rate $\eta_{\alpha} = 3 \times 10^{-4}$, momentum $\beta = (0.5, 0.999)$ and weight decay $10^{-3}$.

\noindent
\textbf{Training setting.}
To make the comparison fair, we follow work~\cite{dwivedi2020benchmarking} for training procedure (data splits, optimizer, metrics, Etc.) and structure (batch normalization, residual connection, Etc.). Specifically, we use Adam optimizer~\cite{maclaurin2015gradient} with the same learning rate decay strategy for all models. An initial learning rate is selected in $\{10^{-3}, 10^{-4}\}$, which is reduced by half if the validation loss does not improve after a fixed number of epochs, either $5$ or $10$. Considering that the network's depth has a significant impact on performance, we compare different methods at a fixed depth. Besides, edge features are excluded since not all methods can take advantage of edge features. We run each experiment with $4$ different seeds. 

\subsection{Results on node classification task} 
\label{sec:nc}
As discussed in work~\cite{dwivedi2020benchmarking}, the PATTERN dataset tests the fundamental graph task of recognizing specific predetermined subgraphs~\cite{scarselli2008graph} and the CLUSTER dataset aims at identifying community clusters in a semi-supervised setting~\cite{kipf2016semi}, where structural information on graph matters. The experimental results are reported in Table \ref{tab_nc}. We have the following observations; first, GIN (with sum aggregator) performs superiority over GCN (with mean aggregator), GraphSage (with max aggregator), and MLP (without considering graph topology) on PATTERN and CLUSTER datasets. This proves that the sum aggregator does better in capturing structural information on the graph better than mean and max aggregators. 
Second, our GNAS achieves significant performance improvement compared to traditional GNNs. 
Third, our GNAS also performs better than the current RL-based method GraphNAS. This benefits from our novel-designed search space from GAP. In contrast, GraphNAS uses a coarse-grained search space with existing GNNs as atomic operations. Further, GNAS has fewer parameters and trains faster than GraphNAS~\cite{gao2020graph}. 

\begin{figure}[tp]
    \centering
    \includegraphics[scale=0.50, trim = 0 0 0 0,clip]{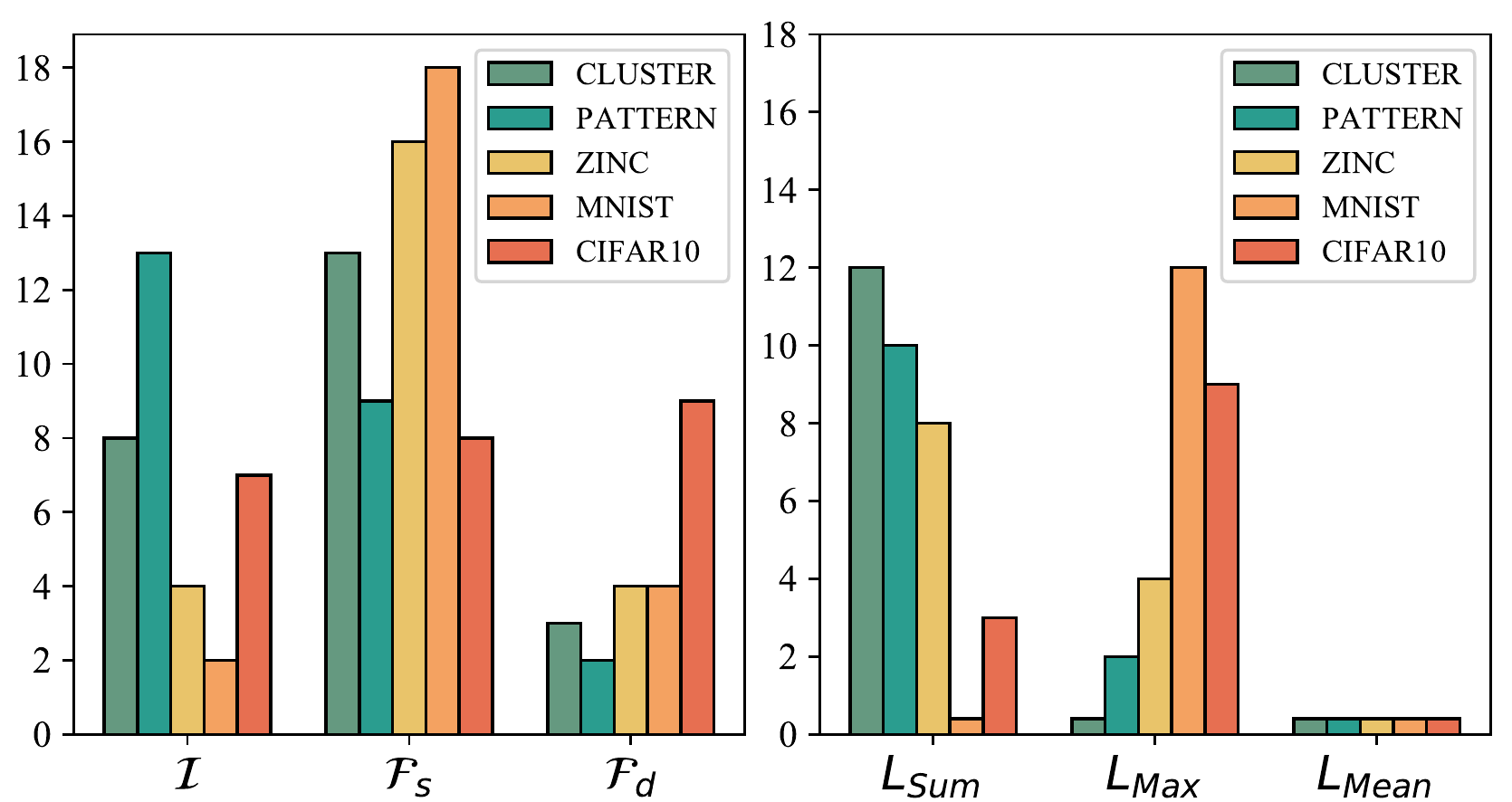}
    \caption{The distribution of searched operations about feature filtering (left figure) and neighbor aggregation (right figure) on five datasets. }
    \label{fig_distribution}
\end{figure}

Besides, we analyze the operations distribution of network searched by GNAS. We find that our GNAS automatically selects the optimal operations to build graph neural architecture for each dataset. As illustrated in Figure~\ref{fig_distribution}, the sum aggregator dominates the distribution of neighbor aggregation. For feature filtering, we find that the selection frequency of dense feature filter~($\mathcal{F}_{d}$) is significantly lower than that of sparse feature filter~($\mathcal{F}_{s}$) on PATTERN and CLUSTER datasets because the original node features are extremely simple on both datasets. 

\subsection{Results on graph classification task} 
The super-pixels datasets test graph classification using the popular MNIST and CIFAR10 image classification datasets, which embeds the ``skeleton''~(super-pixel) of the object into a graph.  
Table~\ref{tab_gc} shows the comparision results. 
Similar observations to node classification tasks~(Section~\ref{sec:nc}) are obtained, e.g., our GNAS also achieves higher performance than traditional and search-based GNNs on both datasets. 
Besides, we also find that (1) max aggregator has the strength to recognize the ``skeleton'' of an object on the graph and ignore the noise nodes. 
The proof is that GraphSage achieves consistent performance improvement over GCN and GIN. This is also found in \cite{xu2018how}. 
(2) MLP model without considering graph topology even performs better than some GNN models, which means the structural information is dispensable. 
(3) From the perspective of operations distribution~(Figure~\ref{fig_distribution}), GNAS prefers selecting max aggregator than sum and mean aggregators for constructing final graph architecture. Further, the dense filter $\mathcal{F}_{d}$ is selected more frequently on CIFAR10 than MNIST since the node features are more involved in CIFAR10.

\subsection{Results on graph regression task} 
ZINC dataset tests the task of graph property regression for contrained solubility, a vital chemical property for designing generative GNNs for molecules. 
Table~\ref{tab_gr} reports the comparision of different methods. 
First, we observe that the performance of GNAS surpasses all the traditional GNN and SOTA GrpahNAS. 
Furthermore, even the MAE of GNAS with 4-Depth is better than other approaches with 16-Depth. 
We attribute this to the efficient message processing capability explored by GNAS. 
Second, the experiments verify the conclusion in literatures~\cite{li2018deeper,rong2019dropedge,Wu_2020} that GNNs' performance cannot always be increased by stacking more layers. 
Third, our GNAS jointly selects sum and max aggregators when searching, where the sum aggregator captures structural information, and the max aggregator focuses on the representative node. 


\begin{table}[tbp]
    \small
    \setlength{\tabcolsep}{2.0 mm}
    \renewcommand\arraystretch{1.2}
    \begin{tabular}{c|c|c|c|c|c}
    \hline
    \multicolumn{2}{c|}{}              & \multicolumn{4}{c}{GRAPH REGRESSION-ZINC} \\ \hline
    Model                                                       & Depth & \#Params & MAE  & Search & Train      \\ \hline \hline
    MLP                                                         & 4     & $0.10$M  & $0.706$     & - & 0.03 hr      \\ 
    \multirow{2}{*}{GCN~\cite{kipf2016semi}}                    & 4     & $0.10$M  & $0.459$     & - & 0.16 hr     \\ 
                                                                & 16    & $0.50$M  & $0.367$     & - & 0.71 hr   \\ 
    \multirow{2}{*}{GIN~\cite{xu2018how}}                       & 4     & $0.10$M  & $0.387$     & - & 0.10 hr     \\ 
                                                                & 16    & $0.50$M  & $0.526$     & - & 0.42 hr     \\ 
    \multirow{2}{*}{GraphSage~\cite{hamilton2017inductive}}     & 4     & $0.10$M  & $0.468$     & - & 0.15 hr     \\ 
                                                                & 16    & $0.50$M  & $0.398$     & - & 0.68 hr     \\ 
    \multirow{2}{*}{GAT~\cite{velickovic2018graph}}             & 4     & $0.10$M  & $0.475$     & - & 0.11 hr     \\ 
                                                                & 16    & $0.53$M  & $0.384$     & - & 0.53 hr     \\ 
    \multirow{2}{*}{GatedGCN~\cite{bresson2017residual}}        & 4     & $0.10$M  & $0.435$     & - & 0.28 hr     \\ 
                                                                & 16    & $0.41$M  & $0.340$     & - & 0.96 hr     \\ 
    \multirow{2}{*}{MoNet~\cite{monti2017geometric}}            & 4     & $0.11$M  & $0.397$     & - & 0.10 hr     \\ 
                                                                & 16    & $0.50$M  & $0.292$     & - & 0.52 hr     \\ 
    \multirow{4}{*}{GraphNAS~\cite{gao2020graph}}               & 4     & $0.48$M  & $0.480$     & 120 hr & 0.45 hr     \\ 
                                                                & 8     & $1.07$M  & $0.413$     & 120 hr & 0.88 hr     \\ 
                                                                & 12    & $1.67$M  & $0.492$     & 120 hr & 1.20 hr     \\ 
                                                                & 16    & $2.23$M  & $0.540$     & 120 hr & 1.66 hr     \\ \hline
    \multirow{4}{*}{\textbf{GNAS}}                              & 4     & $0.41$M  & $0.276$     & 0.35 hr & 0.20 hr     \\ 
                                                                & 8     & $0.82$M  & $0.266$     & 0.72 hr & 0.39 hr     \\ 
                                                                & 12    & $1.20$M  & $\textbf{0.242}$ & 1.10 hr & 0.56 hr  \\ 
                                                                & 16    & $1.68$M  & $0.260$     & 1.75 hr & 0.82 hr      \\ \hline
    \end{tabular}
    \vspace{0.5em}
    \caption{Results of the model searched by our GNAS in comparision with \emph{state-of-the-art} methods on graph regression task, including number of parameters, MAE metric, searching cost and training cost. Lower MAE indicates better performance. }
    \label{tab_gr}
    \vspace{-0.5cm}
 \end{table}

 \subsection{Disscussion of message-passing depth}
\begin{figure}[tp]
    \centering
    \includegraphics[scale=0.48, trim = 0 0 0 0,clip]{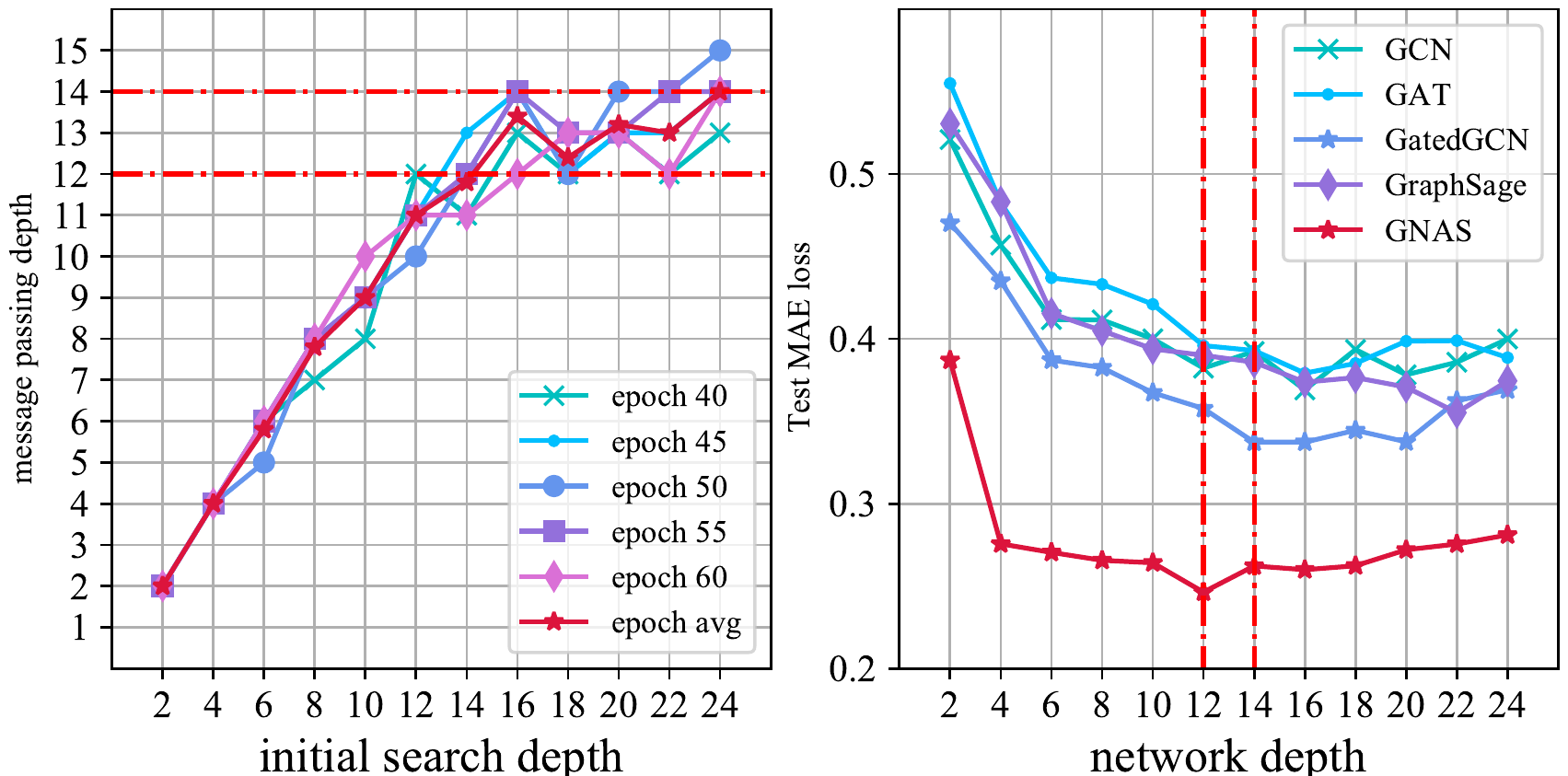}
    \caption{Message-passing depth searched by GNAS at different initial search depth (left), and performance of GNNs at different network depth (right) on ZINC dataset. }
    \label{fig_depth1}
 \end{figure}

As illustrated in Figure \ref{fig_depth1}, we have conducted the experiments with different initial search depths on ZINC dataset. We observe that when the initial depth is less than 12, the searched message-passing depth increases with initial depth. Continuing to increase the initial depth, and the searched depth converges to within the range of 12-14. This indicates that the optimal message-passing depth on the ZINC dataset is between 12 and 14. To verify this, we evaluate the performance of the common GNNs with different message-passing depth. The depth parameter is set to a range of 2-24 with an interval of 2. As shown in the right panel of Figure \ref{fig_depth1}, when the depth of the network is between 12 and 14, the network performance is approximately optimal. This depth range is consistent with that searched by GNAS. 
This demonstrates that our GNAS has the capability for learning optimal message-passing depth. 
We can also find that the performance of the architectures searched by GNAS is far better than that of the manual designed GNNs.

\section{Conclusion}
In this paper, we study NAS for GNN from the message-passing mechanism. 
A graph neural architecture paradigm~(GAP) is designed with two types of atomic operations and tree-topology computation procedure. 
Based on this paradigm, we propose GNAS with a three-level search space and an efficient gradient-based search strategy.  
GNAS can search for better graph architectures with optimal message-passing depth, which has been the focus of researchers' attention in the graph domain. 
Experiment results on five datasets at three fundamental graph tasks demonstrate that GNAS surpasses all human-made and search-based GNNs.

\textbf{Acknowledgement.} This work was supported in part by the National Key R\&D Program of China under Grand:2018AAA0102003, in part by National Natural Science Foundation of China: 61771457, 61732007, 61772494, U19B2038, and in part by the Fundamental Research Funds for the Central Universities.
{\small
\bibliographystyle{ieee_fullname}
\bibliography{egbib}
}

\clearpage
\appendix

\begin{figure*}[ht]
    \centering
    \includegraphics[scale=0.44, trim = 0 0 0 0,clip]{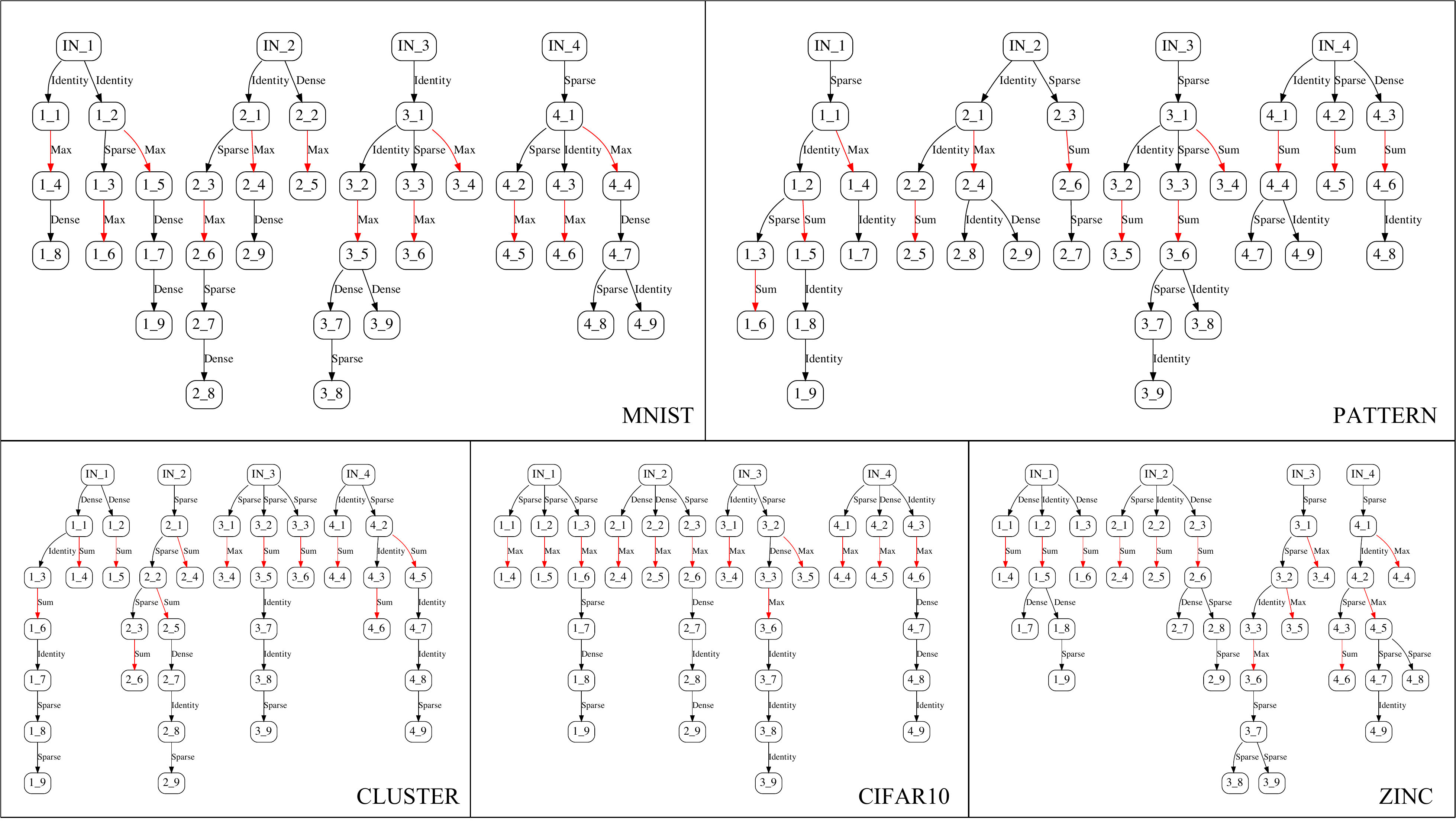}
    \caption{Illustration of graph architecture searched by our GNAS on datasets~(MNIST, PATTERN, CLUSTER, CIFAR10 and ZINC). ``Identity'', ``Sparse'', and ``Dense'' are feature filters. ``Max'', ``Sum'', and ``Mean'' are neighbor aggregators. Node ``Input\_x'' is the input graph embedding of $x$-th graph architecture layer. Node ``x\_y'' denotes the $y$-th latent graph embedding in $x$-th graph architecture layer. }
    \label{fig_searched}
\end{figure*}

\section{Derivation}
In this section, we derive that the existing popular GNNs (GCN~\cite{kipf2016semi}, GIN~\cite{xu2018how}, GraphSage~\cite{hamilton2017inductive}, GAT~\cite{velickovic2018graph}, and GatedGCN~\cite{bresson2017residual}) can be represented using our proposed Graph Architecture Paradigm (GAP). 

\noindent
\textbf{Graph Convolutional Network (GCN)}. In the simplest formulation of GNNs, Graph ConvNet updates node features via an averaging operation over the neighborhood node features, i.e.
\begin{equation}
    h'_{i} = \sigma(\frac{1}{deg_i}\sum_{j \in \mathcal{N}(i)} \textbf{W}^T h_{j}),
\end{equation}
where $\sigma(\cdot)$ is ReLU function, $\textbf{W} \in \mathbb{R}^{d \times d}$ is the learnable parameter, $deg_i$ denotes the degree of node $i$. Let's rewrite it in terms of matrix operations: 
\begin{equation}
    \textbf{H}_{out} = \sigma(L_{mean}(\textbf{H}_{in})\textbf{W}),
\end{equation}
Using $\mathcal{M}(\cdot)$ to replace $\sigma(\cdot)$ and $\textbf{W}$, it can be further approximated as
\begin{equation}
    \textbf{H}_{out} \approx \mathcal{M}(L_{mean}(\textbf{H}_{in})).
\end{equation}

\noindent
\textbf{Graph Isomorphism Network (GIN)}. The GIN architecture is based the Weisfeiler-Lehman Isomorphism Test to study the expressive power of GNNs. The node update equation is defined as
\begin{equation}
    h'_{i} = \mathcal{M}((1+\epsilon)h_i + \sum_{j \in \mathcal{N}(i)} h_j)
\end{equation}
where $\epsilon$ is a learnable constant, $\mathcal{N}(\cdot)$ denotes the set of node neighbors. Note that, $\epsilon$ is a scaling factor for embeddings while sparse filter $\mathcal{F}_s(\cdot)$ achieves better. We replace $\epsilon$ with $\mathcal{F}_s(\cdot)$ and rewrite the equation in terms of matrix operations:
\begin{equation}
    \textbf{H}_{out} \approx \mathcal{M}([\mathcal{I}(\textbf{H}_{in}) || \mathcal{F}_s(\textbf{H}_{in})||L_{sum}(\textbf{H}_{in})]),
\end{equation}

\noindent
\textbf{GraphSage}. GraphSage improves upon the simple GCN model by explicitly incorporating each node’s own features. Using mean aggregator, its update equation is written as
\begin{equation}
    h'_{i} = \sigma (\textbf{U}^T[h_{i} \parallel mean_{j \in \mathcal{N}(i)}(h_{j})])
\end{equation}
where $\textbf{U},\textbf{V} \in \mathbb{R}^{d \times d}$, $\sigma(\cdot)$ denotes ReLU function. We use $\mathcal{M}(\cdot)$ to approximate the equation and rewrite it in terms of matrix operations: 
\begin{equation}
    \textbf{H}_{out} \approx \mathcal{M}([\mathcal{I}(\textbf{H}_{in}) \parallel L_{mean}(\textbf{H}_{in})]).
\end{equation}

\noindent
\textbf{Graph Attention Network (GAT)}. GAT uses the attention mechanism to introduce anisotropy in the neighborhood aggregation function. The node update equation is given by:
\begin{equation}
    h'_{i} = \sigma(\sum_{j \in \mathcal{N}(i)}\textbf{W}^T \alpha_{ij} h_j)
\end{equation}
where $\textbf{W} \in \mathbb{R}^{d \times d}$ is a learnable parameter. $\alpha_{ij} \in \mathbb{R}$ denotes the similarity between $h_i$ and $h_j$, formulated as 
\begin{equation}
    \alpha_{ij} = \mathcal{S}(h_i,h_j)
\end{equation}
where $\mathcal{S}(\cdot, \cdot)$ is the function that compute similarity for two features. We use two $\mathcal{M}(\cdot)$ to approximate $\mathcal{S}(\cdot,\cdot)$:
\begin{equation}
    \alpha_{ij} \approx \mathcal{M}_1(h_i)\mathcal{M}_2(h_j), 
\end{equation}
where the output of $\mathcal{M}(\cdot)$ is a constant.
The node update equation can be approximated by
\begin{equation}
    h'_{i} \approx \sigma(\textbf{W}^{T}\mathcal{M}_1(h_i)\sum_{j \in \mathcal{N}(i)}\mathcal{M}_2(h_j)h_j).
\end{equation}
We rewrite it in terms of matrix operation:
\begin{equation}
    \textbf{H}_{out} \approx \sigma ( \mathcal{M}_1(\textbf{H}_{in})L_{sum}(\mathcal{M}_2(\textbf{H}_{in})\textbf{H}_{in})\textbf{W}).
\end{equation}
Note that, $\mathcal{M}(\textbf{H}_{in})$ can be viewed as scaling factor, which can be extended to sparse filter $\mathcal{F}_s(\cdot)$. Therefore, the formula can be further approximated as
\begin{equation}
    \textbf{H}_{out} \approx \mathcal{M}(\mathcal{F}_s(L_{sum}(\mathcal{F}_s(\textbf{H}_{in})))).
\end{equation}

\noindent
\textbf{Gated Graph ConvNet(GatedGCN)}. GatedGCN considers edge gates to design another anisotropic variant of GCN. The authors propose to explicitly update edge features along with node features:
\begin{equation}
    h' = \sigma(\textbf{U}^Th_i+\sum_{j \in \mathcal{N}(i)}\eta_{ij}\odot \textbf{V}^Th_j),
\end{equation}
where $\textbf{U},\textbf{V} \in \mathbb{R}^{d\times d}$, $\odot$ denotes hardmard product, $\sigma(\cdot)$ is ReLU function. Edge gate $\eta_{ij} \in \mathbb{R}^{d}$ is defined as
\begin{equation}
    \eta_{ij} = \sigma(\textbf{A}^Th_i + \textbf{B}^Th_j),
\end{equation}
where $A, B \in \mathbb{R}^{d \times d}$, $\sigma(\cdot)$ denotes sigmoid function. 
Similar with the approximation of GAT, we use $\mathcal{M}(\cdot)$ to approximate $\eta_{ij}$:
\begin{equation}
    \eta_{ij} \approx \mathcal{M}_1(h_i) \odot \mathcal{M}_2(h_j)
\end{equation}
where the output of $\mathcal{M}(\cdot)$ is a vector in $\mathbb{R}^{d}$. We rewrite the update formula in terms of matrix operations:
\begin{equation}
    \textbf{H}_{out} \approx \sigma(\textbf{H}_{in}\textbf{U} + (\mathcal{M}_1(\textbf{H}_{in})\odot \sum_{j \in \mathcal{N}(i)}\mathcal{M}_2(\textbf{H}_{in})\odot\textbf{H}_{in})\textbf{V})
\end{equation}
Note that, $\mathcal{M}(\textbf{H}_{in})$ can be viewed as dense scaling factor, which can be extended to dense filter $\mathcal{F}_d(\cdot)$. We further replace the outermost parameters and activation functions with $\mathcal{M}(\cdot)$:
\begin{equation}
    \textbf{H}_{out} \approx \mathcal{M}([\mathcal{I}(\textbf{H}_{in})\parallel \mathcal{F}_d(L_{sum}(\mathcal{F}_d(\textbf{H}_{in})))])
\end{equation}

\section{Searched Graph Architectures}

As shown in Figure~\ref{fig_searched}, we visualize the graph architectures searched by our GNAS on five datasets. 
We find that first GNAS prefers ``Max'' aggregator at graph classification task~(on both MNIST and CIFAR10 datasets), because ``Max'' can capture ``skeleton'' of object on graph. Second, the searched graph architectures at graph classification task~(on both PATTERN and CLUSTER datasets) contain lots of ``Sum'' aggregators. This results from that ``Sum'' aggregator can better capture structural information~\cite{xu2018how}, where structural information matters on PATTERN and CLUSTER datasets. Third, ``Mean'' aggregator is not selected by GNAS when constructing the graph architectures on all datasets, which demonstrates the mean statistic dose not work on these datasets. On ZINC dataset, we focus on molecular property prediction, both structural information and representative atom have significant influence on the prediction results. Therefore, GNAS selects both ``Sum'' and ``Max'' aggregators to capture corresponding information. 

\begin{figure*}[tp]
    \centering
    \includegraphics[scale=0.65, trim = 0 0 0 0,clip]{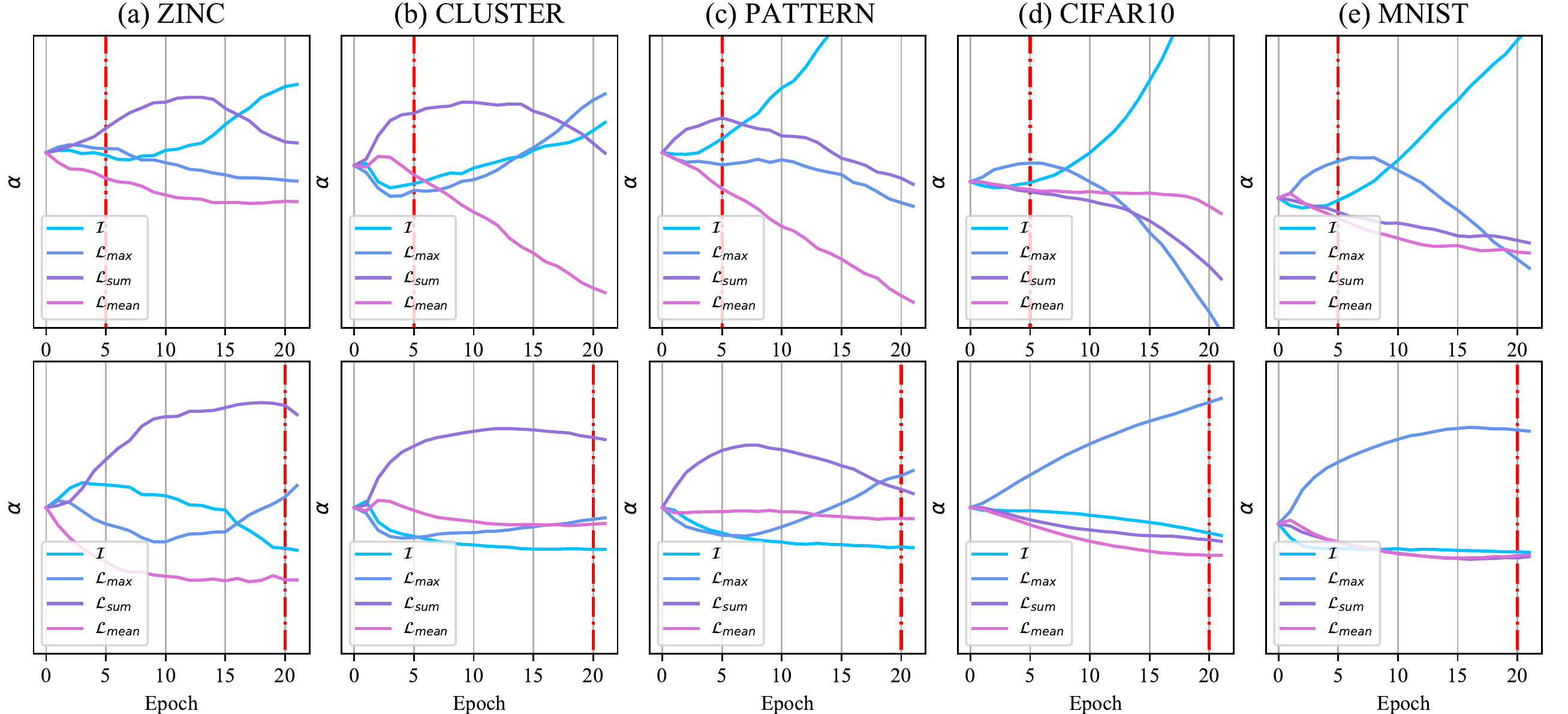}
    \caption{The distribution of neighbor aggregation operations optimized by both training loss and validation loss (top row), and only by training loss (bottom row). }
    \label{fig_earlystopping}
\end{figure*} 

\begin{table*}[t]
    \vspace{0.5cm}
    \small
    \setlength{\tabcolsep}{3.7 mm}
    \renewcommand\arraystretch{1.2}
    \begin{tabular}{c|c|c|c|c|c|c|c|c|c}
    \hline
    \multicolumn{2}{c|}{}              & \multicolumn{8}{c}{NODE CLASSIFICATION}                                      \\ \hline
    \multicolumn{2}{c|}{}              & \multicolumn{4}{c|}{PATTERN}          & \multicolumn{4}{c}{CLUSTER}          \\ \hline
    Model                      & Depth & Param & Test Acc & Train Acc & Time   & Param & Test Acc & Train Acc & Time   \\ \hline \hline
    GCN                        & 16    & 0.50M & 71.89    & 78.41     & 11.3hr & 0.50M & 68.51    & 71.73     & 6.08hr \\ 
    GIN                        & 16    & 0.51M & 85.39    & 85.66     & 0.62hr & 0.52M & 64.72    & 65.97     & 0.47hr \\ 
    GraphSage                  & 16    & 0.50M & 50.49    & 50.49     & 5.19hr & 0.50M & 63.84    & 86.71     & 3.70hr \\ 
    GAT                        & 16    & 0.53M & 78.27    & 90.21     & 0.77hr & 0.53M & 70.56    & 76.07     & 0.75hr \\ 
    GatedGCN                   & 16    & 0.50M & 85.57    & 86.01     & 11.9hr & 0.50M & 73.84    & 87.88     & 6.81hr \\ 
    MoNet                      & 16    & 0.51M & 85.58    & 85.72     & 1.58hr & 0.51M & 66.41    & 67.73     & 1.05hr \\ \hline 
    \textbf{GNAS}              & 16    & 1.60M & \textbf{86.85}    & 86.69     & 3.52hr & 1.60M & \textbf{74.77}    & 75.02     & 3.21hr \\ \hline
    \end{tabular}
    \vspace{0.0em}
    \caption{Extra experimental results at node classification task. }
    \label{tab:ext}
\end{table*}

\section{Collapse problem}

For differentiable search strategies of DARTS~\cite{liu2018darts}, there exists the ``collapse'' problem. 
This is discussed in work~\cite{liang2019darts+}: after certain searching epochs, the number of identity operation increases dramatically in the selected architecture, which results in poor performance of the selected architecture. 
We visualize the operations distribution searched by GNAS on five datasets in the top row of Figure~\ref{fig_earlystopping}, and find that GNAS also suffers from this problem. 
Although ``early stopping'' trick~\cite{liang2019darts+} can alleviate this phenomenon, it makes the search for the graph neural architecture insufficient. 
Here, we attempt to unify the optimization objectives of weights $w$ and architecture parameters $\alpha$ to minimize the training loss without considering the validation loss. Surprisingly, the ``collapse'' problem disappears. 
As shown in the bottom row of Figure~\ref{fig_earlystopping}, the number of identity operation no longer increases with searching epochs, so it brings GNAS sufficient optimization without ``earlystopping''. 
We attribute this to the consistency in the objectives of optimizing weights $w$ and architecture $\alpha$. 

In the original method, the architecture parameters $\alpha$ are optimized by validation loss, and the weights $w$ are optimized by training loss, whose optimization objectives are inconsistent in the later stages. Since the validation loss is very close to training loss at the beginning of the search procedure, $w$ and $\alpha$ are optimized towards the same objective. As the model begins to overfit the training set, the gap between the validation loss and the training loss widens. The data that feeds to the last cells are not conducive to reducing validation loss. Therefore, the last cells attend to select more identity operations to obtain good feature representations directly from the previous cells.


\section{Supplementary Experimental Results}
We conduct the experiments at node classification task~(on both PATTERN and CLUSTER datasets) and report the results in Table~\ref{tab:ext}. Considering that GraphNAS~\cite{gao2020graph} takes too long to search at depth 16, we ignored it in the comparison. From the comparision, we can find that GNAS can always achieve the best performance compared to traditional GNNs no matter the message passing depth is 4 or 16. 

\end{document}